\documentclass[10pt,twocolumn,letterpaper]{article}

\usepackage[pagenumbers]{cvpr}
\usepackage{pifont}

\definecolor{linkblue}{rgb}{0.21,0.49,0.74}
\usepackage[breaklinks,colorlinks,allcolors=linkblue]{hyperref}

\title{Revisit with Evidence: Agentic Video Generation with Spatial Memory}

\author{
  Qirui Zhong$^{1}$ \quad Juncheng Wang$^{2}$ \quad Shujun Wang$^{2}$ \quad Cheng Xue$^{1}$\textsuperscript{*}\\
  $^{1}$Southeast University \quad $^{2}$Hong Kong Polytechnic University\\
  {\tt\small qiruizhong@seu.edu.cn}
}

\date{}

\makeatletter
\g@addto@macro\@maketitle{%
  \begin{center}
    \begin{minipage}{\textwidth}
      \centering
      \includegraphics[width=\linewidth]{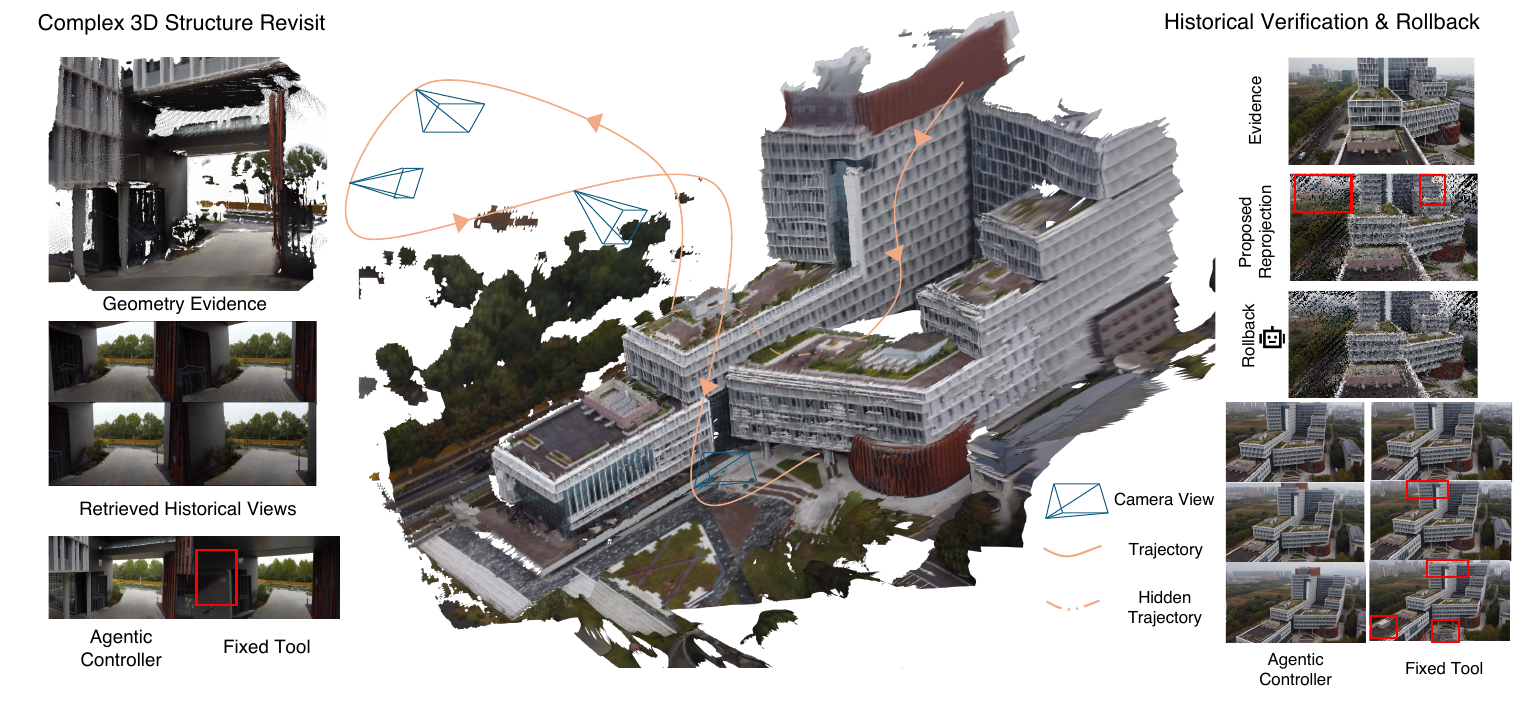}
      \captionsetup{hypcap=false}
      \captionof{figure}{\textbf{Evidence-guided revisiting under occlusion.}
      Left: the controller retrieves historical observations and invokes
      Memory and Geometry tools for an occluded passage, with a fixed-tool
      result for comparison. Right: historical verification rejects a
      proposed scene revision and triggers rollback to the accepted
      estimate; the generated views below compare the RL agent with the
      Fixed-tool baseline using fixed Memory and Geometry gains. The
      central scene and illustrative paths provide spatial context.}
      \label{fig:teaser}
    \end{minipage}
  \end{center}
}
\makeatother

\begin{document}
\maketitle

\begingroup
\renewcommand{\thefootnote}{*}
\footnotetext{Corresponding author: Cheng Xue.}
\endgroup

\begin{abstract}
Camera-controlled video generation should preserve scenes across
revisits despite incomplete historical evidence. We introduce
\textbf{Revisit-with-Evidence}, an agentic framework that learns to
coordinate visual Memory and Geometry tools around a frozen video
backbone. An evidence-conditioned planner selects historical
references, conditioning strengths, and tool calls using observation
support and tool feedback. After imitation initialization,
trajectory-level reinforcement learning on real generation rollouts
optimizes video quality, task completion, and execution cost;
checkpoints enable continued training from additional real
interactions. Immutable observations remain distinct from scene
estimates and generated hypotheses. A fixed verifier checks revisions
against current and fixed initial references, returning feedback for
replanning and rollback. Across 160 DL3DV development captures,
the trained RL agent reduces LPIPS from 0.3403 to 0.3336 and
improves the VBench six-dimension quality subscore by 2.84 points
over Fixed-tool Top4 using the same conditioning modules.

\end{abstract}

\section{Introduction}
\label{sec:intro}

Camera-controlled video generation requires spatial consistency across
viewpoint changes: when a camera revisits a location, previously observed
objects should retain their appearance and spatial arrangement.
This capability supports interactive games, virtual walkthroughs, and
embodied navigation simulators.
Maintaining this consistency remains challenging when previously observed
regions leave the generator's bounded temporal
context~\cite{xiao2026worldmem,wu2026video}.

Visual and spatial memory methods retain historical evidence to improve
consistency across revisits~\cite{xiao2026worldmem,li2025vmem,wu2026video}.
Existing approaches preserve historical appearance using image or latent memories~\cite{xiao2026worldmem,zhang2025storymem}, while estimated geometry supports view retrieval or provides rendered spatial conditions~\cite{li2025vmem,wu2026video,lee20263d}. These mechanisms offer complementary cues about appearance and spatial structure, but the evidence available for a requested view can be incomplete or unreliable. View overlap may fail to account for occlusion~\cite{yu2025context,xiao2026worldmem}, and reconstruction or accumulated generation errors can compromise the conditions used for subsequent synthesis~\cite{wu2026video,yang2026geometry}. A central question is therefore how to select and combine visual and geometric evidence according to its visibility, relevance, and reliability.

We introduce \textbf{Revisit-with-Evidence}, an agentic framework that
learns to coordinate visual Memory and Geometry tools around a frozen
camera-conditioned video generator. We formulate reference selection,
conditioning strength, and tool scheduling as a sequential decision
problem, and train an evidence-conditioned planner through
trajectory-level reinforcement learning on real video-generation
rollouts. The policy uses historical support, regional conflicts, and
tool feedback to choose evidence and actions, optimizing a return that
balances video quality, task completion, and computational cost.
It can be further trained as new scene data become available.
An observation-grounded verifier preserves real observations and checks
proposed scene revisions against both the current accepted state and a
fixed initial reference. Rejected revisions trigger rollback, while
generated hypotheses retain their predicted status.
Figure~\ref{fig:teaser} illustrates conditioning under occlusion (left)
and scene verification with rollback (right).
Our contributions are threefold:
\begin{itemize}
    \item An evidence-conditioned planning policy trained with trajectory-level reinforcement learning to jointly select historical references, configure Memory and Geometry gains, and schedule tool use. The policy supports continued optimization from additional real interactions.
    \item An observation-grounded verification and scene-update mechanism that distinguishes real observations, geometric estimates, and generated hypotheses, checks revisions against current and fixed initial references, and supplies feedback for replanning and rollback.
    \item Experiments on DL3DV~\cite{ling2024dl3dv} combining system comparisons and conditioning ablations with real-trajectory policy-training and continuation studies, documenting both the complementary roles of visual and geometric evidence and the policy update process.
\end{itemize}

\section{Related Work}
\label{sec:related_work}

\subsection{Interactive Video Generation}
\label{sec:rw_interactive_generation}

Interactive video generation supports user-directed actions and camera
motion. Genie~3 enables real-time world navigation~\cite{parker2025genie},
while GameNGen predicts gameplay frames from visual history and
actions~\cite{valevski2025diffusion}. WorldPlay reconstitutes historical
context for long-term geometric consistency~\cite{sun2025worldplay}.
PhysVid uses temporally localized, physics-aware
conditioning~\cite{pathak2026physvid}.

Motion and subject control provide further customization. Wan-Move guides object motion using latent guidance derived from dense point trajectories~\cite{chu2026wan}, whereas VideoComposer integrates textual, spatial, and temporal conditions through a unified spatio-temporal condition encoder~\cite{wang2023videocomposer}. SMRABooth aligns subject and motion representations with the video model during LoRA fine-tuning, enabling joint customization of appearance and motion~\cite{xu2026smrabooth}.

For explicit viewpoint control, CameraCtrl introduces a camera-pose conditioning module for video diffusion models~\cite{he2025cameractrl}, whereas SCoPE incorporates camera rays into positional encoding~\cite{yin2026scope}. Geometry-conditioned approaches further use scene structure to guide camera-controlled synthesis~\cite{lee20263d,huang2026cinescene}. We adopt the camera-conditioned generation setting.


\subsection{Memory-Conditioned Video Generation}
\label{sec:rw_memory_generation}

Visual memory methods retain historical appearance in images or latent
representations. WorldMem associates memory frames with poses and
timestamps, using state-aware attention to recover content across
viewpoint and temporal gaps~\cite{xiao2026worldmem}. StoryMem targets
cross-shot storytelling through a compact keyframe bank maintained by
semantic selection, aesthetic filtering, and long-term memory
anchors~\cite{zhang2025storymem}.

Geometry structures memory retrieval and conditioning. Context-as-Memory
retrieves frames by camera-FOV overlap~\cite{yu2025context}, while VMem
indexes views through visible surfels~\cite{li2025vmem}. SpMem combines
static point-cloud conditioning with recent and episodic
frames~\cite{wu2026video}. GEAR uses occlusion-filtered patch
correspondences for memory attention~\cite{yang2026geometry}, and
Lyra~2.0 uses geometry to retrieve frames and establish dense
correspondences~\cite{shen2026lyra}. Spatia updates persistent point-cloud
memory through visual SLAM~\cite{zhao2026spatia}.

Autoregressive errors can degrade subsequent conditioning. Self Forcing
trains on self-generated contexts to reduce exposure
bias~\cite{huang2026self}. Memory Forcing uses chained predicted rollouts
in Minecraft~\cite{huang2025memory}, while GEAR and Lyra~2.0 train with
degraded or self-augmented
histories~\cite{yang2026geometry,shen2026lyra}. We learn evidence selection
and conditioning strengths while validating proposed scene updates.

\begin{figure*}[t]
  \centering
  \includegraphics[width=\textwidth]{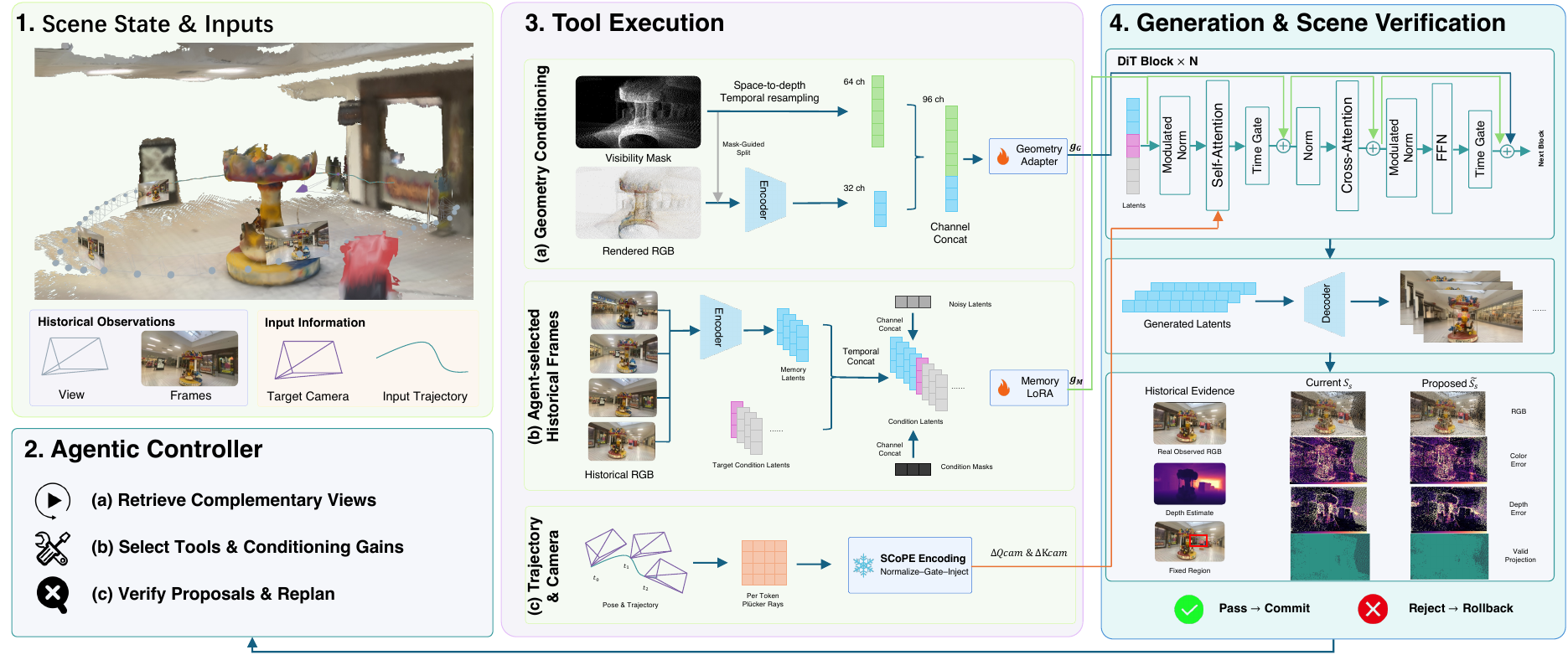}
  \caption{\textbf{Framework overview.}
  (1) Historical observations, scene estimates, and target cameras form
  the inputs. (2) An RL planner selects references, gains, and tool calls.
  (3) Memory, Geometry, and SCoPE provide appearance, structure, and
  camera conditioning. (4) The frozen generator produces candidates;
  a fixed verifier governs scene commitment and rollback.
  Tool feedback drives replanning.}
  \label{fig:overview}
\end{figure*}

\subsection{Agentic Video Generation}
\label{sec:rw_agentic_generation}

Agentic video generation coordinates tools and feedback around
generators. NEWTON learns physics-aware tool planning and replanning
with Flow-GRPO~\cite{feng2026newton}. VideoGen-Agent combines supervised
fine-tuning and multitask reinforcement learning for augmentation,
generation, and verification~\cite{li2026videogen}.

For multi-shot continuity, MVAgent constructs viewpoint-specific
references and transition conditions, optimizing orchestration through
shot-level candidate comparisons~\cite{kong2026mvagent}. ATAS maintains
registered geometry and time-indexed event memory, checking reconstructed
observations before state admission~\cite{wang2026any}. Complementary RL
post-training targets video world models directly: WorldCompass uses
clip-level rollouts and rewards for interaction accuracy and visual
quality~\cite{wang2026worldcompass},
while WorldCycle uses reversible action cycles for spatial and temporal
consistency rewards~\cite{gu2026worldcycle}.

Our work focuses on learning evidence-conditioned control for
camera-controlled scene revisiting. We jointly optimize historical
reference selection, visual and geometric conditioning strengths, and
tool scheduling from real generation trajectories around a frozen
generator. Historical verification constrains scene revisions and
returns feedback to the policy, linking learned orchestration to reliable
and persistent spatial evidence.

\section{Methodology}
\label{sec:methodology}

\subsection{Problem Formulation and Overview}
\label{sec:problem_formulation}

%
\paragraph{Problem formulation.}
Given historical observations $\mathcal{H}=\{(I_i,T_i,K_i)\}_{i=1}^{N}$,
where $T_i$ and $K_i$ are camera-to-world poses and intrinsics, we
generate an $F$-frame clip $\widehat{\mathcal{V}}_k$ conditioned on an initial frame
$I_k^0$, a text prompt $p_k$, and a target camera sequence
$\mathcal{C}_k=\{(T_{k,f},K_{k,f})\}_{f=0}^{F-1}$.
The goal is to follow the requested trajectory while preserving the
appearance and layout of observed scene regions. Subsequent clips start
from the final frame of the preceding accepted clip.
Target RGB references are accessed only after rollout sealing, for
policy-training rewards or evaluation; they never enter deployment
observations.

\paragraph{Controllable generation interface.}
Our Wan2.2/SCoPE backbone~\cite{yin2026scope} comprises a causal 3D VAE,
a diffusion transformer (DiT), and camera-ray encoding. Two conditioning
plugins expose historical appearance and estimated geometry as
independently controllable inputs:
\begin{equation}
  \begin{aligned}
    \widehat{\mathcal{V}}_k
      &= \mathcal{G}_{\theta,\phi_M,\phi_G}
         (I_k^0,p_k,\mathcal{C}_k;\mathcal{U}_k,\xi_k),\\
    \mathcal{U}_k
      &= (\mathcal{R}_k,\mathcal{Q}_k,g_k^M,g_k^G),
  \end{aligned}
  \label{eq:generation_interface}
\end{equation}
where $\mathcal{R}_k\subseteq\mathcal{H}$ contains selected historical
views, $\mathcal{Q}_k$ contains target-view renderings of the accepted
scene estimate and their visibility masks, and $\xi_k$ is sampling noise.
The tool arguments $\mathcal{U}_k$
specify both the evidence supplied to the generator and its conditioning
strengths.

\paragraph{Plugin injection.}
Following StoryMem~\cite{zhang2025storymem}, the Memory path incorporates
independently encoded historical images as latent memory slots. Its LoRA
adapters modify projections within self-attention, cross-attention, and
feed-forward layers. The Geometry branch adds conditioning residuals to
target-token features at selected DiT blocks. At an adapted linear layer
and a geometry injection site, respectively,
\begin{equation}
  \begin{aligned}
    \operatorname{Linear}_M(x)
      &= Wx + g_k^M\frac{\alpha}{\sqrt{r}}BAx,\\
    h_\ell &\leftarrow h_\ell + g_k^G\Delta h_\ell^G,
  \end{aligned}
  \label{eq:plugin_injection}
\end{equation}
where $W$ is a frozen projection, $A,B$ are rank-$r$ adapter matrices,
and $\alpha$ is the adapter scale. At block $\ell$, $h_\ell$ denotes
target-token features and $\Delta h_\ell^G$ the geometry-branch residual
conditioned on $\mathcal{Q}_k$.
Disabling Memory removes both its latent slots and LoRA residuals;
disabling Geometry bypasses its residual branch. The plugins
$\phi_M,\phi_G$ are trained offline with the host $\theta$ frozen;
all network weights remain fixed during agent execution.

\paragraph{Agentic execution.}

At step $s$, the planner in Fig.~\ref{fig:overview} uses the accepted
scene $S_s$, observations $\mathcal{H}$, target cameras $\mathcal{C}_k$,
tool history $\mathcal{L}_s$, and remaining budget $b_s$:
\begin{equation}
  \begin{aligned}
    o_s &= f(S_s,\mathcal{H},\mathcal{C}_k,\mathcal{L}_s,b_s),\\
    a_s &\sim \pi_\psi(\cdot\mid o_s,\mathcal{A}_s).
  \end{aligned}
  \label{eq:agentic_interface}
\end{equation}
The observation $o_s$ summarizes evidence support, coverage, conflicts,
task progress, and tool feedback. The policy $\pi_\psi$ selects reference
sets, Memory/Geometry gains, tool calls and regions, or termination
from the admissible set $\mathcal{A}_s$, whose candidates satisfy evidence
eligibility, execution prerequisites, and budget limits. Generation actions specify
$\mathcal{U}_k$ in Eq.~\eqref{eq:generation_interface}; tool receipts
update $\mathcal{L}_s$ for replanning.

We initialize $\psi$ by imitating the rule policy and optimize it on real
generation rollouts using trajectory-level GRPO-style updates, keeping
the backbone and plugins frozen. Deployment fixes $\psi$ and executes
the highest-probability admissible action.
Sections~\ref{sec:spatial_memory_conditioning}--\ref{sec:verified_scene_updates}
detail conditioning, planning, and verification; Supplementary Sec.~D
covers learning and continuation.

\subsection{Spatial Memory and Conditioning}
\label{sec:spatial_memory_conditioning}

\paragraph{Spatial memory and observation support.}
We maintain immutable historical RGB images and calibrated cameras in
$\mathcal{H}$ and a revisable scene estimate $S_s$.
Frozen VGGT-$\Omega$~\cite{wang2026vggt} reconstructs $S_0$ from
$\mathcal{H}$. Surfaces retain links to supporting observations, with
separate provenance for estimated geometry and generated hypotheses.

For a surface element $p$ in $S_s$, we project it into historical
views and evaluate its support using visibility, depth and color
consistency, and reconstruction confidence. Let $w_i^{(s)}(p)$ denote
the resulting support score from observation $i$, with zero support
assigned to invalid or occluded correspondences. We define
\begin{equation}
  \begin{aligned}
    \mathcal{O}_s(p)
      &= \{i : w_i^{(s)}(p)>0\},\\
    q_s(p)
      &= \mathbf{1}[|\mathcal{O}_s(p)|\geq 2],
  \end{aligned}
  \label{eq:observation_support}
\end{equation}
where $\mathcal{O}_s(p)$ indexes distinct real observations and
$q_s(p)$ indicates sufficient independent support. Generated views
do not increase the observed support count. These support relations
provide coverage and conflict summaries for the policy observation
$o_s$ and help construct admissible historical reference choices in
$\mathcal{A}_s$ (Sec.~\ref{sec:agentic_control}).

\paragraph{Condition construction and adaptation.}
The Geometry path in Fig.~\ref{fig:overview} (3a) renders $S_s$ along
$\mathcal{C}_k$ to form $\mathcal{Q}_k$. RGB renderings undergo mask-guided
preparation and VAE encoding, while visibility masks undergo spatial
packing and temporal alignment to the latent grid. Their features are
concatenated along the channel dimension for the Geometry adapter,
which supplies the residuals in Eq.~\eqref{eq:plugin_injection}.

In panel (3b), up to four selected historical images in $\mathcal{R}_k$
are independently VAE-encoded and temporally concatenated with
target-condition latents constructed from the initial frame and padding.
Together with noisy latents
and condition masks, they form the channel-concatenated input to patch
embedding. The resulting visual tokens are processed by the
LoRA-adapted DiT described in Sec.~\ref{sec:problem_formulation}.
Panel (3c) converts target cameras into per-token Pl\"ucker rays;
the frozen SCoPE module injects their encoded features into
self-attention queries and keys. These conditioning paths meet in
the generator shown in stage (4).

Plugin adaptation follows two stages. We first train Geometry while
freezing the host and Memory weights, then fine-tune only the Memory
LoRA with Geometry frozen and alternately enabled or bypassed.
This adapts Memory for use with and without geometric conditioning.
The adapted plugins remain frozen during policy learning and deployment.
At inference, the trained policy $\pi_\psi$ selects the historical
reference set $\mathcal{R}_k$ and conditioning gains $(g_k^M,g_k^G)$
from admissible candidates, as described in Sec.~\ref{sec:agentic_control}.

\subsection{RL for Evidence-Guided Planning}
\label{sec:agentic_control}

The controller turns the evidence representation in
Sec.~\ref{sec:spatial_memory_conditioning} into decisions about reference
selection, conditioning strengths, and tool execution. We instantiate
Eq.~\eqref{eq:agentic_interface} with a learned agentic planner
$\pi_\psi$. Each tool returns observations and an execution receipt,
which enter $\mathcal{L}_s$ before the next decision. The planner learns
to coordinate these decisions from real generation trajectories and
their task returns.

\paragraph{Diagnosing the available evidence.}
At each tool step, the controller refreshes historical support on the
current scene $S_s$. It records source quality and confidence, surface
coverage, and regional depth and color disagreement with real history.
The diagnostics distinguish insufficient support from a contradiction
backed by distinct real observations. These summaries enter $o_s$ so
that the policy can decide whether to gather evidence, change its
conditioning, or attempt a supported revision. Generated images and
their reconstructions retain their predicted status throughout this
process.

\paragraph{Selecting complementary historical views.}
A fixed Memory budget makes redundant views costly. Let
$\mathcal{P}_s$ be the sampled scene surface and $\mathcal{A}$ the views
already included during candidate construction. Using the support
scores from Sec.~\ref{sec:spatial_memory_conditioning}, the marginal
evidence supplied by view $i$ is
\begin{equation}
  \Delta_s(i\mid\mathcal{A})=
  \sum_{p\in\mathcal{P}_s}\omega_s(p)
  \big[w_i^{(s)}(p)-c_{\mathcal{A}}^{(s)}(p)\big]_+,
  \label{eq:marginal_evidence_gain}
\end{equation}
where $[x]_+=\max(x,0)$, $c_{\mathcal{A}}^{(s)}(p)$ is the support
already covered, and $\omega_s(p)$ weights target visibility and
diagnosed coverage deficits. Covered support starts at zero and is
refreshed after each addition, so later candidates reflect the evidence
they add. Source quality and available confidence further describe the
reliability of each view. These quantities construct and characterize
eligible candidates. Reference-set alternatives combine marginal
support, quality, coverage, and supported observation anchors; the
learned policy selects among them and retains at most four views as
$\mathcal{R}_k$. Thus, complementary coverage remains explicit, while
the final reference-set choice is made by $\pi_\psi$.

\paragraph{Controlling conditioning strengths.}
Memory and Geometry provide different evidence: selected historical
images supply appearance, while the accepted scene estimate supplies
spatially aligned structure. Their reliability changes with reference
quality, observed support, coverage, and regional conflicts. The planner
uses these diagnostics to choose an admissible pair $(g_k^M,g_k^G)$
from discrete conditioning configurations. The gains control the
Memory LoRA and Geometry residuals through
Eq.~\eqref{eq:plugin_injection}; an empty reference set bypasses Memory,
and conservative configurations are available under unknown feedback.
The gain decision is learned jointly with reference selection and tool
scheduling. The original continuous gain rules remain documented as
the rule-controller baseline in Supplementary Sec.~B.4.

\paragraph{Learning the agentic planner.}
We parameterize $\pi_\psi$ with a compact differentiable tree that
scores admissible candidates using observation and action features.
Training begins with imitation of rule-controller decisions from the
training split, followed by trajectory-level GRPO-style
updates~\cite{shao2024deepseekmath} on real tool interactions and
generated videos. The return balances reference fidelity, temporal
consistency, task completion, and execution costs. Rollouts sharing an
initial context receive group-normalized return advantages, which are
assigned to all decisions in each trajectory. Clipped policy updates
with a fixed-reference KL penalty optimize the planning policy.
Target RGB provides reward labels only after rollout sealing.
Additional real trajectories support continued optimization by
restoring the policy, reference, and optimizer states; the generator
and conditioning plugins remain frozen. Supplementary Sec.~D gives the
policy, reward, and optimization formulations.

\paragraph{Replanning after tool feedback.}
Before generation, the policy can query up to eight additional
available historical views when considering insufficient or conflicting
evidence. Returned observations refresh the diagnostics and available
reference sets. A generation action instantiates $\mathcal{U}_k$ in
Eq.~\eqref{eq:generation_interface} and produces a predicted candidate.
After generation, reconstruction can combine that candidate with
queried real views, and repair can construct a reversible scene
proposal. The policy schedules these calls subject to their execution
prerequisites and the remaining budget. A proposal must pass the fixed
observation-grounded verifier before commitment; failed checks trigger
rollback (Sec.~\ref{sec:verified_scene_updates}). Tool receipts and
validation outcomes enter $\mathcal{L}_s$, informing subsequent queries,
generation, or termination. Video acceptance and scene commitment remain
separate, and subsequent clips start from the accepted clip's endpoint.
During deployment, $\psi$ is fixed and the highest-probability
admissible action is executed; the episode ends upon completion, a stop
decision, or budget exhaustion.

\begin{figure}[!t]
  \centering
  \includegraphics[width=\columnwidth]{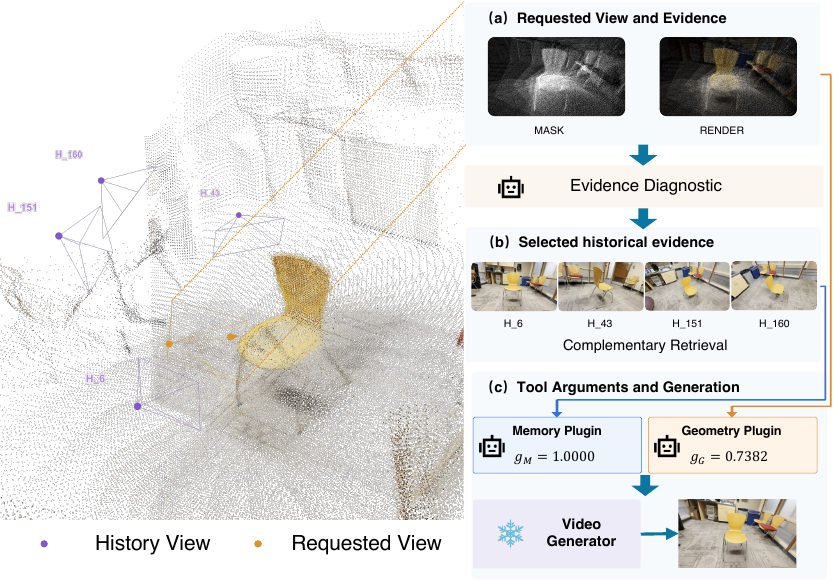}
  \caption{\textbf{RL agent generation example.}
  Left: selected historical cameras (purple) and the requested camera
  (orange) in the scene estimate. Right: (a) geometry rendering and
  visibility mask; (b) historical observations $H_6,H_{43},H_{151},H_{160}$;
  (c) RL-selected gains $g_k^M=1.0000$, $g_k^G=0.7382$, plugin conditioning,
  and a generated frame.}
  \label{fig:controller_case}
\end{figure}

\paragraph{Example execution.}
Figure~\ref{fig:controller_case} illustrates an execution of the trained
RL agent. The
observations $H_6,H_{43},H_{151},H_{160}$ view the yellow chair from
different directions and provide historical appearance, while the scene
rendering and visibility mask provide spatial conditioning. The learned
policy $\pi_\psi$ selects historical references, conditioning gains,
and tool calls. The recorded gains configure both plugins around the
frozen generator.

\subsection{Verified State Updates}
\label{sec:verified_scene_updates}

Scene revisions affect subsequent evidence retrieval and geometric
conditioning. We therefore validate each proposed update before it
enters the persistent scene state. The learned planner schedules
verification, while the fixed SceneGuard determines whether a proposal
can be committed. Its admission criteria remain frozen during policy
training, continuation, and deployment.

\paragraph{Source-bound proposals.}
A proposal $\widetilde{S}_s$ records its parent scene version,
affected regions, and supporting historical observations.
Corrections to estimated geometry require support from distinct real
observations. Generated images and reconstructed hypotheses retain
their predicted provenance, while the original observations
$\mathcal{H}$ remain immutable.

\paragraph{Validation against current and initial references.}
Figure~\ref{fig:overview} (4) visualizes depth error $E_d$, color error
$E_c$, and coverage $C$ on fixed historical reference pixels, using
real RGB observations and depth estimated from those observations.

Depth and color errors must not increase, and coverage must not
decrease, relative to both $S_s$ and $S_0$. These checks apply globally
and within each evaluable affected region. The current reference
constrains individual revisions, while the fixed initial reference
limits cumulative degradation. Supplementary Sec.~C.4 gives the full
inequalities and numerical tolerance.

\paragraph{Commitment and rollback.}
Only proposals with valid provenance and passing applicable checks
are committed. Rejection discards the uncommitted proposal and retains
the preceding accepted scene. Committed updates affect subsequent
retrieval and rendering. Video-candidate acceptance and scene commitment
are separate: accepting a generated clip does not automatically admit
its reconstructed geometry into the scene state. Generated hypotheses
retain their predicted status after commitment.

\paragraph{Feedback to the planner.}
Verification receipts add global and regional errors, coverage,
provenance violations, and rejection reasons to $\mathcal{L}_s$,
informing subsequent reference selection, conditioning gains, and tool
calls. The resulting transitions are retained for policy training.

Figure~\ref{fig:teaser} (right) provides an example: verification
detects a regional depth-error regression in the proposed revision and
triggers rollback. The restored scene is identical to the preceding
accepted estimate.

\begin{table*}[t]
  \centering
  \caption{Comparison on 160 DL3DV development captures with three seeds and
  three segments per seed (1,440 accepted clips per method). Scores are averaged within each
  capture and then equally across captures. VBench reports its six-dimension
  quality total and individual dimension scores, all on a 0--100 scale.
  Best and second-best values are \textbf{bold} and
  \underline{underlined}, respectively; ties share the same highlighting.
  Ours uses the trained RL planner, frozen during evaluation.}
  \label{tab:main_comparison}
  \footnotesize
  \setlength{\tabcolsep}{3pt}
  \renewcommand{\arraystretch}{1.18}
  \begin{tabular*}{\textwidth}{@{\extracolsep{\fill}}*{12}{c}@{}}
    \toprule
    & & \multicolumn{3}{c}{Image Quality}
      & \multicolumn{7}{c}{VBench $\uparrow$} \\
    \cmidrule(lr){3-5}\cmidrule(l){6-12}
    Method & \shortstack{Camera\\input}
      & LPIPS $\downarrow$ & PSNR $\uparrow$ & SSIM $\uparrow$
      & \shortstack{Total\\(6D)}
      & \shortstack{Subject\\consist.}
      & \shortstack{Background\\consist.}
      & \shortstack{Aesthetic\\quality}
      & \shortstack{Imaging\\quality}
      & \shortstack{Motion\\smooth.}
      & \shortstack{Dynamic\\degree} \\
    \midrule
    VMem (native)~\cite{li2025vmem} & Yes & 0.6997 & 11.098 & 0.5218 & 65.37 & 68.56 & 85.40 & 37.28 & 51.04 & 93.29 & \textbf{100.00} \\
    VMem (history-4)~\cite{li2025vmem} & Yes & 0.6433 & 10.772 & 0.4972 & 68.73 & 74.42 & 86.93 & 48.68 & 51.41 & 92.64 & \textbf{100.00} \\
    SpMem (adapted)~\cite{wu2026video} & Yes & 0.7029 & 11.493 & 0.5516 & 66.55 & 77.10 & 88.34 & 36.38 & 36.40 & 97.45 & 87.50 \\
    StoryMem~\cite{zhang2025storymem} & No & 0.6407 & 10.904 & 0.4840 & 75.15 & 86.10 & 91.08 & 50.08 & 54.83 & 96.50 & \underline{95.83} \\
    WorldMem (diagnostic)~\cite{xiao2026worldmem} & No & 0.7886 & 7.843 & 0.3358 & 68.74 & 80.20 & \underline{91.67} & 30.11 & 45.80 & \underline{97.79} & 86.67 \\
    \midrule
    Fixed-tool Top4 & Yes & \underline{0.3403} & \underline{15.810} & \underline{0.6306} & \underline{75.84} & \underline{86.52} & 90.82 & \underline{51.81} & \underline{55.06} & 96.39 & \textbf{100.00} \\
    \textbf{Ours (full agent)} & Yes & \textbf{0.3336} & \textbf{15.987} & \textbf{0.6318} & \textbf{78.68} & \textbf{88.15} & \textbf{91.91} & \textbf{52.03} & \textbf{60.92} & \textbf{98.18} & \textbf{100.00} \\
    \bottomrule
  \end{tabular*}
  \par\vspace{3pt}
  \begin{minipage}{\textwidth}
    \footnotesize
    Reference metrics use a common field of view and exclude the initial
    frame; VBench uses complete native clips. Total (6D) is the normalized,
    weighted quality subscore of the six reported dimensions. Camera input
    denotes an explicit target trajectory. SpMem uses camera and
    sequence-length adaptations; WorldMem is a cross-domain diagnostic
    using native Minecraft no-op actions. Native model settings and
    compute budgets differ across methods.
  \end{minipage}
\end{table*}

\section{Experiments}
\label{sec:experiments}

\subsection{Experimental Setup}
\label{sec:experimental_setup}
\paragraph{Datasets and evaluation protocol.}
We conduct experiments on real-world captures from
DL3DV-10K~\cite{ling2024dl3dv}. The conditioning modules are adapted
on 32 training captures using scene-reconstruction supervision.
Our main development comparison covers 160 captures, three random
seeds, and three commanded segments.
Each segment contains 65 frames, and subsequent segments start
from the endpoint of the preceding accepted clip.
The controller accesses up to 299 historical observations, with
four historical views reserved for validation and excluded from
retrieval. For planner training, target RGB references are accessed
after rollout sealing to compute rewards; during evaluation, they are
used only for scoring. They never enter policy observations.
Controller ablations use 12 development captures and three random seeds,
with one candidate per segment.

Module ablations use 12 development captures. Fixed-tool Top4 in
Table~\ref{tab:main_comparison} provides the fixed-conditioning
reference, detailed in Supplementary Sec.~E.1.

All experiments are conducted on NVIDIA H100 80GB GPUs.

\paragraph{Implementation details.}
We use Wan2.2-I2V-A14B~\cite{wan2025wan} with SCoPE camera
conditioning~\cite{yin2026scope} as the frozen backbone.
We first train the Geometry branch and then adapt the
StoryMem-based Memory LoRA~\cite{zhang2025storymem} with Geometry
alternately enabled and disabled. The two stages use AdamW
with learning rates of $10^{-5}$ and $2\times10^{-6}$,
respectively. All network weights remain fixed during evaluation.
Our method generates videos at $832\times480$ resolution and
16 fps using 40 sampling steps, a guidance scale of 3.5,
and a sampling shift of 5. The controller uses at most four
Memory slots and can request up to eight additional historical
views per query.

\paragraph{Planner training.}
The policy-learning study uses a single-segment development protocol
with one training capture and a separate frozen-policy evaluation
capture. Only planner parameters are optimized; the backbone,
conditioning plugins, and SceneGuard criteria remain fixed. Training
and continuation details are provided in Supplementary Sec.~D.

\paragraph{Compared methods.}
Our principal in-house baseline uses the same backbone,
conditioning weights, observed history, initial geometry,
and paired sampling seeds as the full agent. It selects four
historical views using independent spatial scores and fixes
the Memory and Geometry gains to 1.0 and 0.75, respectively.
This baseline generates one candidate per segment, whereas
the full agent permits at most two candidates.

The system comparison in Table~\ref{tab:main_comparison} and the
controller ablations in Table~\ref{tab:controller_ablation} evaluate
the trained RL planner, with policy parameters fixed throughout
evaluation.
We additionally evaluate VMem~\cite{li2025vmem}, including a
four-history-view initialization variant,
SpMem~\cite{wu2026video}, and StoryMem~\cite{zhang2025storymem}.
SpMem uses our declared camera and sequence-length adaptations;
StoryMem retains its native memory interface without the same
explicit camera-trajectory input.
WorldMem~\cite{xiao2026worldmem} is included as a cross-domain
diagnostic reference. Method-specific inference settings and
adaptations are reported separately.

\paragraph{Evaluation metrics.}
We measure reference fidelity using LPIPS, PSNR, and SSIM
under a common spatial field of view, excluding the initial
conditioning frame. For the planner study, this exclusion follows the
source-frame clock and removes all positions associated with the known
initial observation. Perceptual video quality is evaluated
using a six-dimension VBench~\cite{huang2024vbench} quality subscore covering
subject consistency, background consistency, aesthetic quality,
imaging quality, motion smoothness, and dynamic degree.
VBench uses the complete native clips, including the conditioning
frame, with its own preprocessing.
We first average repeated measurements within each capture
and then report an equally weighted mean across captures.
WorldScore evaluates world-generation controllability, quality, and
dynamics~\cite{duan2025worldscore}; R2M-Bench compares revisit pairs with
gap-matched non-revisit and short-range controls from the same
rollout~\cite{gu2026r2m}. These benchmarks motivate
complementary diagnostics for camera adherence and revisit-specific
memory beyond the fidelity and video-quality scores reported here.
The system-comparison results are summarized in
Table~\ref{tab:main_comparison}.

\subsection{Main Results}
\label{sec:main_results}

The trained RL agent reduces LPIPS from 0.3403 to 0.3336 (an absolute
decrease of 0.0067) and raises VB-6 from 75.84 to 78.68 (+2.84 points) over
Fixed-tool Top4 (Table~\ref{tab:main_comparison}).
The system comparison uses up to two candidates per segment for the agent
and one for Fixed-tool, while Table~\ref{tab:controller_ablation} assesses
controller components under a shared $K=1$ budget.
These results support learned tool coordination for improved reference
fidelity and perceptual video quality in this evaluation.

We also resumed planner training from a saved checkpoint using
additional real generation rollouts (Supplementary Sec.~D.5).

\begin{figure}[!t]
  \centering
  \includegraphics[width=\columnwidth]{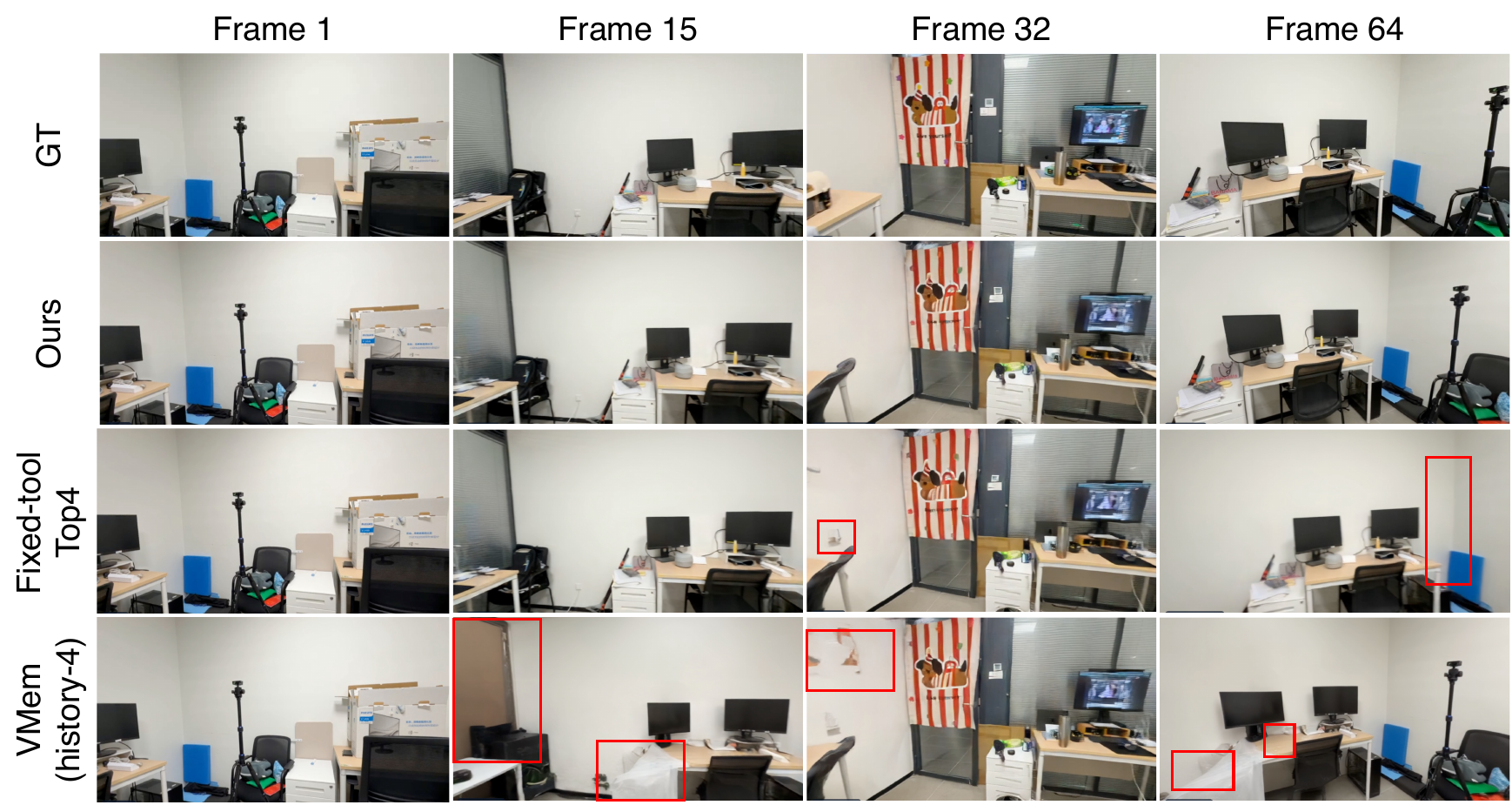}
  \caption{\textbf{Scene Consistency on Revisits.}
  Rows show GT, Ours, Fixed-tool Top4, and VMem (history-4);
  columns show frames 1, 15, 32, and 64. Red boxes mark missing
  or altered scene details in the baselines.}
  \label{fig:qualitative_comparison}
\end{figure}

\subsection{Ablation Studies}
\label{sec:ablation_studies}

We evaluate conditioning complementarity (Table~\ref{tab:module_ablation})
and controller components under $K=1$ (Table~\ref{tab:controller_ablation}).

\begin{table}[htbp]
  \centering
  \caption{Memory--Geometry ablation on 12 development captures.
  M: Memory; G: Geometry; VB-6: VBench six-dimension quality.}
  \label{tab:module_ablation}
  \small
  \setlength{\tabcolsep}{4pt}
  \renewcommand{\arraystretch}{1.15}
  \begin{tabular*}{\columnwidth}{@{\extracolsep{\fill}}*{6}{c}@{}}
    \toprule
    M & G & LPIPS $\downarrow$ & PSNR $\uparrow$
      & SSIM $\uparrow$ & VB-6 $\uparrow$ \\
    \midrule
                 &              & 0.5213 & 12.054 & 0.5295 & 75.64\\
    $\checkmark$ &              & 0.4677 & 12.702 & 0.5387 & 75.31\\
                 & $\checkmark$ & \underline{0.3768} & \underline{15.416}
      & \underline{0.6074} & \underline{75.69}  \\
    $\checkmark$ & $\checkmark$ & \textbf{0.2680} & \textbf{16.505}
      & \textbf{0.6349} & \textbf{75.92} \\
    \bottomrule
  \end{tabular*}
\end{table}

\paragraph{Complementarity of Memory and Geometry.}

Under the separate reconstruction protocol, either path improves reference
fidelity; combining Memory's appearance and Geometry's spatial structure
gives the best LPIPS and VB-6 (Table~\ref{tab:module_ablation}).
Weights remain frozen; enabled Geometry uses gain 1.

\paragraph{Controller components.}

Table~\ref{tab:controller_ablation} uses 12 development captures, three
paired seeds, and three segments per seed ($K=1$), with both plugins
and the settings and aggregation of Sec.~\ref{sec:experimental_setup}.
Disabling R uses independent spatial Top4; disabling G fixes
$(g^M,g^G)=(1.0,0.75)$; disabling Q removes extra queries; disabling U
retains $S_0$.
Updates follow Sec.~\ref{sec:verified_scene_updates}.
This protocol is separate from the system comparison; policies remain frozen.

\begin{table}[htbp]
  \centering
  \caption{RL controller ablation on 12 development captures with
  three random seeds ($K=1$). R: complementary retrieval;
  G: gain control; Q: extra query; U: verified scene update.}
  \label{tab:controller_ablation}
  \small
  \setlength{\tabcolsep}{3pt}
  \renewcommand{\arraystretch}{1.15}
  \begin{tabular*}{\columnwidth}{@{\extracolsep{\fill}}*{7}{c}@{}}
    \toprule
    Variant & R & G & Q & U & LPIPS $\downarrow$ & VB-6 $\uparrow$ \\
    \midrule
    Fixed Top4       & \ding{55} & \ding{55} & \ding{55} & \ding{55}
      & 0.4270 & 75.25 \\
    w/o retrieval    & \ding{55} & \ding{51} & \ding{51} & \ding{51}
      & 0.4120 & 76.42 \\
    w/o gain control & \ding{51} & \ding{55} & \ding{51} & \ding{51}
      & \underline{0.3801} & 76.91 \\
    w/o extra query  & \ding{51} & \ding{51} & \ding{55} & \ding{51}
      & 0.3828 & \underline{77.46} \\
    w/o scene update & \ding{51} & \ding{51} & \ding{51} & \ding{55}
      & 0.3852 & 77.08 \\
    \midrule
    \textbf{Full ($K=1$)} & \ding{51} & \ding{51} & \ding{51} & \ding{51}
      & \textbf{0.3548} & \textbf{78.28} \\
    \bottomrule
  \end{tabular*}
\end{table}

Full ($K=1$) reduces LPIPS from 0.4270 to 0.3548 (16.91\%) and raises VB-6
from 75.25 to 78.28 (+3.03 points) over Fixed Top4.
Each removal worsens both metrics; removing retrieval most degrades LPIPS.
All variants generate once per segment, excluding additional candidate sampling.

\begin{figure}[!t]
  \centering
  \includegraphics[width=\columnwidth]{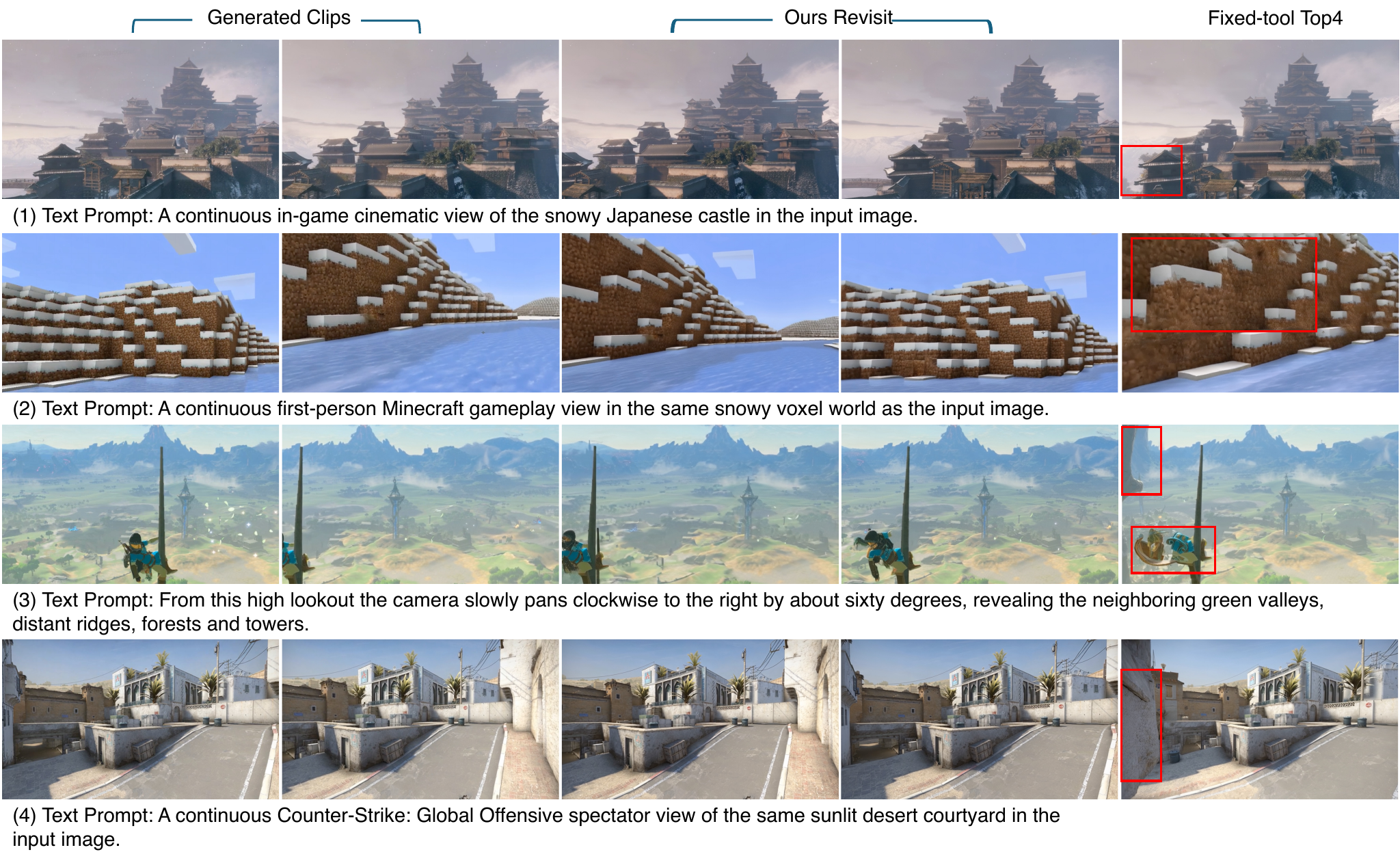}
  \caption{\textbf{Open-Domain Generation.} Rows show Sekiro, Minecraft, Zelda, and CS:GO. Columns 1–2 show exploration, columns 3–4 show our generated return, and column 5 shows the Fixed-tool Top 4 return endpoint. Red boxes highlight deviations from the initial scene.}
  \label{fig:open_domain}
\end{figure}

\subsection{Qualitative Results}
\label{sec:qualitative_results}
Figure~\ref{fig:qualitative_comparison} compares matched indoor revisit
keyframes. Fixed-tool Top4 omits the slender tripod at frame 64;
VMem (history-4) alters the reflective window at frame 15; both add
patterns to the plain wall at frame 32. Geometry can miss thin
structures and cannot fully specify view-dependent reflections;
mismatched references may transfer unrelated textures. Our method
better preserves these details, supporting complementary appearance
and geometry through reference selection and gain control.

Figure~\ref{fig:open_domain} shows single-screenshot exploration and
revisiting across four games with realistic, voxel, and cel-shaded
styles beyond DL3DV. After exploration, we independently generate a
return along the reversed estimated camera path. Generated history
remains predicted; all weights remain fixed without game-specific training.

\section{Conclusion}
\label{sec:conclusion}

We presented \textbf{Revisit-with-Evidence}, which learns reference
selection, Memory/Geometry gains, and tool use around a frozen
camera-conditioned generator through reinforcement learning on real
generation rollouts for spatially consistent revisiting.
A fixed verifier checks revisions against current and fixed initial
references, preserving immutable observations and hypothesis provenance
while informing replanning and rollback.
Restoring policy and optimizer checkpoints enables continued training from
additional real interactions.
Historical coverage and reconstruction quality remain limiting factors,
motivating future work on sparse observations and longer revisit sequences.

\clearpage
\section*{LLM Usage Statement}
We used OpenAI Codex to assist with experimental-code implementation
and debugging, and with drafting and language revision across the
manuscript. Codex also supported preliminary
discussions of experimental protocols and exploratory diagram drafts.
The authors determined the final experimental design and prepared the
final figures.

{
    \small
    \bibliographystyle{ieeenat_fullname}
    \bibliography{main}
}

\end{document}